\documentclass[review]{elsarticle}
\usepackage{hyperref}

\usepackage{graphics}
\usepackage{graphicx}
\newsavebox\CBox

\usepackage{amsmath,amssymb,amsfonts, commath}
\usepackage{tabularx}
\usepackage{subcaption} 
\usepackage{booktabs}
\usepackage{siunitx}
\usepackage[capitalise]{cleveref}
\crefformat{appendix}{#2#1#3}
\usepackage[symbol]{footmisc}
\usepackage{diagbox}
\usepackage{xspace}
\usepackage{longtable}
\usepackage{float}

\DeclareMathOperator*{\E}{\mathbb{E}}
\newcommand{\minus}{\scalebox{0.75}[1.0]{$-$}}

\newcommand{\panomaly}{p_m^{anomaly}(\textbf{x}^*)}

\newcommand{\WGSS}{$W^{GSS}$\xspace}
\newcommand{\WAUROC}{$W^{AUROC}$\xspace}

\newcommand{\VarNLL}{$\mathrm{Var_{\theta}[NLL]}$\xspace}
\newcommand{\uepi}{$U^{epistemic}$\xspace}
\newcommand{\ualea}{$U^{aleatoric}$\xspace}
\newcommand{\utotal}{$U^{total}$\xspace}
\newcommand{\uexceed}{$U^{exceed}$\xspace}
\newcommand{\GSS}{$GSS$\xspace}

\newcommand{\baseGSS}{$Base^{GSS}$\xspace}
\newcommand{\gainGSS}{$Gain^{GSS}$\xspace}
\newcommand{\NLLfunc}{\ell{}_{\hat{\theta}_m}(}

\newcommand{\qFull}{q_\psi{}(\theta)}
\newcommand{\pFull}{p(\theta|X)}
\newcommand{\qSub}{q_\psi{}(\theta_m)}

\usepackage{pifont}
\newcommand{\cmark}{\ding{51}}%
\newcommand{\xmark}{\ding{55}}%

\usepackage[figurename=Fig.]{caption}

\journal{Expert Systems with Applications}

\begin{document}

\begin{frontmatter}

\title{Bayesian autoencoders with uncertainty quantification: Towards trustworthy anomaly detection}

\author[a1]{Bang Xiang Yong}
\ead{bxy20@cam.ac.uk}

\author[a1]{Alexandra Brintrup}
\ead{ab702@cam.ac.uk}

\address[a1]{Institute for Manufacturing, University of Cambridge, United Kingdom}

\begin{abstract}

Despite numerous studies of deep autoencoders (AEs) for unsupervised anomaly detection, AEs still lack a way to express uncertainty in their predictions, crucial for ensuring safe and trustworthy machine learning systems in high-stake applications. Therefore, in this work, the formulation of Bayesian autoencoders (BAEs) is adopted to quantify the total anomaly uncertainty, comprising epistemic and aleatoric uncertainties. To evaluate the quality of uncertainty, we consider the task of classifying anomalies with the additional option of rejecting predictions of high uncertainty. In addition, we use the accuracy-rejection curve and propose the weighted average accuracy as a performance metric. Our experiments demonstrate the effectiveness of the BAE and total anomaly uncertainty on a set of benchmark datasets and two real datasets for manufacturing: one for condition monitoring, the other for quality inspection. 

\end{abstract}

\begin{keyword}
\texttt{Bayesian autoencoders, anomaly detection, uncertainty, trustworthy machine learning}
\end{keyword}

\end{frontmatter}

\section{Introduction}

The use of unsupervised neural networks (NNs) such as deep autoencoders (AEs) has consistently achieved improvements over traditional machine learning (ML) models for high-dimensional anomaly detection \cite{chalapathy2019deep,munir2019comparative, pang2021}. Nonetheless, one missing ingredient is the measurement of predictive uncertainty. That is, in addition to reporting the prediction of an anomaly, can the AE tell how uncertain it is about the prediction? 

\begin{figure}[H]
\centerline{\includegraphics[scale=.55]{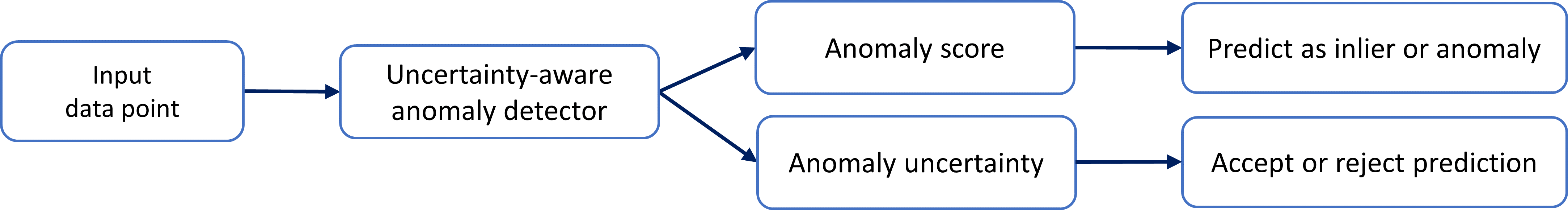}}
\caption{An uncertainty-aware anomaly detector quantifies predictive uncertainty in addition to predicting inliers and anomalies.}
\label{fig:ua-ad}
\end{figure}

The model's prediction and its uncertainty are two different outputs: the prediction classifies a data point as inlier or anomaly by relying on an anomaly score, which typically measures its similarity or distance from a reference distribution of normality; the higher the anomaly score, the higher the chance of an abnormal observation \cite{chandola2009anomaly}. By contrast, the anomaly uncertainty measures the trustworthiness of the classification; the higher the uncertainty, the less reliable is the prediction, implying a greater chance of a predictive error. (See \cref{fig:ua-ad} for an illustration).

One popular method to quantify uncertainty for NNs is by adopting the Bayesian framework, resulting in the Bayesian autoencoder (BAE) formulation. Two types of uncertainty are of interest: epistemic uncertainty captures the uncertainty in the BAE parameters, while aleatoric uncertainty reflects the inherent randomness in the data \cite{kiureghian2009}. However, recent works on BAEs \cite{legrand2019use, yong2020icml, daxberger2019bayesian, tran2021model, baur2020bayesian} have only investigated the use of epistemic and aleatoric uncertainties as anomaly scores, leaving the notion of anomaly uncertainty understudied. Consequently, we are still incapable of knowing when to (dis)trust the BAEs' predictions.

Knowing the uncertainty of each prediction is important for ensuring reliability, especially in high-stake applications, as we can anticipate whether there is an error with the prediction \cite{gao2019towards}. If we filter away cases of high uncertainty, we are left with trustworthy predictions of higher accuracy. In addition, these uncertain cases can be referred for further diagnosis, either by humans or equipment, thereby increasing the overall predictive performance. Therefore, in this work, we contribute in the following ways:

\begin{enumerate}
\item \textbf{Formulation of BAE to quantify predictive uncertainty of anomalies.} We do so by converting negative log-likelihood (NLL) estimates from the BAE into the total anomaly uncertainty which is decomposable into epistemic and aleatoric uncertainties.
\item \textbf{Evaluation of uncertainty in anomaly detection.} We consider the task of classifying anomalies with the option of rejecting uncertain predictions. While these evaluations have been explored in supervised learning \cite{nadeem2009accuracy, hendrickx2021machine, leibig2017leveraging}, we contribute by extending to unsupervised anomaly detection.
\item \textbf{Validation of proposed BAE with real use cases that demonstrate its effectiveness and value.} Our experiments demonstrate the added benefits of quantifying anomaly uncertainty on a set of benchmark datasets, as well as two real datasets for manufacturing applications: one for condition monitoring and the other for quality inspection. 
\end{enumerate}

\section{Background and related works}

Anomaly detection\footnote{Also known as novelty detection, outlier detection, and more recently, out-of-distribution detection.} is a binary classification task of identifying data points that differ significantly from a reference distribution of inliers \cite{chandola2009anomaly}. Anomaly detection has many real use cases in high-stake applications, e.g. cybersecurity \cite{kwon2019survey}, financial surveillance \cite{ahmed2016278}, health and medical diagnosis \cite{fernando2021medical}, manufacturing \cite{kamat2020anomaly}. Many traditional ML models have been adapted to perform anomaly detection, e.g. isolation forest, one-class support vector machine, K-means clustering \cite{agrawal2015708}. A drawback of traditional ML models is their low complexity and flexibility, limiting their scalability from handling high-dimensional data prevalent in modern databases \cite{thudumu2020comprehensive}. 

Overcoming the limitations of traditional ML models, deep NNs are expressive and scalable models. Many works have studied their use for anomaly detection, consistently achieving state-of-the-art results \cite{pang2021,fernando2021medical,kamat2020anomaly,kwon2019survey,ahmed2016278}. A key property of NNs is their flexibility of modelling; each component (e.g. activation function, layers, optimiser) can be tuned and customised to adapt to the data at hand. Furthermore, specialised hardware have expedited their adoption and scalability by parallelising and optimising computations \cite{li2019edge}. 

An emerging challenge with using NNs for anomaly detection is the quantification of uncertainty \cite{thudumu2020comprehensive}. Uncertainty is a core component for promoting algorithmic transparency and hence, the advancement of trustworthy ML \cite{bhatt2021uncertainty}. While most works have focused on anomaly scores, less is known about anomaly uncertainty. To this end, Perini et al. \cite{perini2020quantifying} introduced the example-wise confidence method for quantifying anomaly uncertainty, however, their methods have only experimented with traditional ML models, leaving their applicability to deep AEs unclear. 

The development of Bayesian neural networks (BNNs) \cite{bishop2007pattern} for uncertainty quantification has led to the formulation of BAEs, a probabilistic version of the AE. Nonetheless, extant works have used uncertainty as an anomaly score, which is different from the notion of anomaly uncertainty. We list extant works as follows: Yong et al. \cite{yong2020bayesian} quantified epistemic and aleatoric uncertainties for detecting sensor drifts. Daxberger \& Lobato \cite{daxberger2019bayesian} developed the Bayesian variational autoencoder (BVAE) to quantify uncertainty as a measure of deviation from the training distribution. Baur et al. \cite{baur2020bayesian} developed the BAE with Monte Carlo dropout for anomaly detection in medical application. Tran et al. \cite{tran2021model} observed that BAE-generated data have higher uncertainty than the reconstructed data. Chandra et al. \cite{chandra2021revisiting} studied the BAE with Markov Chain Monte Carlo (MCMC) sampling for dimensionality reduction and classification tasks.

\begin{table}[hbtp]
\centering
\caption{Comparisons between related works and ours on ML models for anomaly uncertainty.}
\label{table:comparison-uncood}
\resizebox{0.7\textwidth}{!}{
    \begin{tabular}{@{}p{0.35\textwidth}cc@{}}
    \toprule
    Work & ML model & {\centering Anomaly uncertainty} \\ \midrule
Ours                & BAE         & \cmark   \\ \midrule
Perini et al. \cite{perini2020quantifying}       & Traditional & \cmark  \\
Yong et al. \cite{yong2020bayesian}               & BAE         & \xmark\\
Daxberger \& Lobato \cite{daxberger2019bayesian} & BVAE         & \xmark  \\
Baur et al. \cite{baur2020bayesian}              & BAE         & \xmark \\
Tran et al. \cite{tran2021model}              & BAE         & \xmark \\
Chandra et al. \cite{chandra2021revisiting}          & BAE         & \xmark  \\
Kriegel et al. \cite{kriegel2011interpreting}        & Traditional & \xmark  \\ 
Kwon et al. \cite{kwon2020uncertainty}      & Supervised BNN & \xmark  \\ \bottomrule
    \end{tabular}
}
\end{table}

Our work stands out as the first to provide a comprehensive study on the BAE's predictive uncertainty of an anomaly. We extend the work of Kwon et al. \cite{kwon2020uncertainty}, which   decomposed the classification uncertainty of supervised BNNs into epistemic and aleatoric uncertainties, to consider unsupervised anomaly detection with BAEs. Further, our work is inspired by Kriegel et al. \cite{kriegel2011interpreting}, who proposed converting an ensemble of anomaly scores into anomaly probabilities, and combining them via averaging. To improve the combination quality, they further proposed a customised scaling. We extend their study by (1) including BAEs, (2) quantifying anomaly uncertainty while maintaining coherence within a probabilistic framework, and (3) finding that their customised scaling improved the quality of uncertainty. Comparisons with related works are summarised in \cref{table:comparison-uncood}.

\section{Methods}
This section details the probabilistic formulation of BAEs for quantifying anomaly uncertainty, followed by ways to evaluate them. 

\subsection{Bayesian autoencoder}

Suppose we have a set of data $X^{train} = {\{\textbf{x}_1,\textbf{x}_2,\textbf{x}_3,...\textbf{x}_N\}}$, $\textbf{x}_{i} \in \rm I\!R^{D}$. An AE is an NN parameterised by $\theta$, and consists of two parts: an encoder $f_\text{encoder}$, for mapping input data $\textbf{x}$ to a latent embedding, and a decoder $f_\text{decoder}$ for mapping the latent embedding to a reconstructed signal of the input $\hat{\textbf{x}}$ (i.e. $\hat{\textbf{x}} = {f_{\theta}}(\textbf{x}) = f_\text{decoder}(f_\text{encoder}(\textbf{x}))$) \cite{goodfellow2016deep}. 

Bayes' rule can be applied to the parameters of the AE to create a BAE,
\begin{equation}\label{eq_posterior}
    p(\theta|X^{train}) = \frac{p(X^{train}|\theta)\  p(\theta)}{p(X^{train})} \\ ,
\end{equation}
where $p(X^{train}|\theta)$ is the likelihood and $p(\theta)$ is the prior distribution of the AE parameters. The log-likelihood for a diagonal Gaussian distribution is,
\begin{equation} \label{eq_gaussian_loss}
\log{p(\textbf{x}|\theta)} = -(\frac{1}{D} \sum^{D}_{i=1}{\frac{1}{2\sigma_i^2}}(x_i-\hat{x_i})^2 + \frac{1}{2}\log{\sigma_i^2})
\end{equation} 
where $\sigma_i^2$ is the variance of the Gaussian distribution. For simplicity, we use an isotropic Gaussian likelihood with $\sigma_i^2=1$ in this study, since the mean-squared error (MSE) function is proportional to the NLL. For the prior distribution, we use an isotropic Gaussian prior which effectively leads to $L_2$ regularisation.

Since Equation~\ref{eq_posterior} is analytically intractable for a deep NN, various approximate methods have been developed such as Stochastic Gradient Markov Chain Monte Carlo (SGHMC) \cite{chen2014stochastic}, Monte Carlo Dropout (MCD) \cite{gal2016dropout}, Bayes by Backprop (BBB) \cite{blundell2015weight}, and anchored ensembling \cite{pearce2020uncertainty} to sample from the posterior distribution. In contrast, a deterministic AE has its parameters estimated using maximum likelihood estimation (MLE) or maximum a posteriori (MAP) when regularisation is introduced. The variational autoencoder (VAE) \cite{kingma2013auto} and BAE are AEs formulated differently within a probabilistic framework: in the VAE, only the latent embedding is stochastic while the $f_\text{encoder}$ and $f_\text{decoder}$ are deterministic and the model is trained using variational inference, while, on the other hand, the BAE (similar to BNN) has distributions over all parameters of $f_\text{encoder}$ and $f_\text{decoder}$. See \cref{app:methods} for descriptions of the posterior sampling methods used in our study.

In short, the training phase of BAE entails using one of the sampling methods to obtain a set of approximate posterior samples $\{\hat{\theta}_m\}^M_{m=1}$. Then, during the prediction phase, we use the posterior samples to compute $M$ estimates of the NLL. For brevity, we denote the posterior NLL score $\minus\log{p(\textbf{x}|\hat{\theta}_m)}$, as $\NLLfunc{\textbf{x})}$. Next, we look into using the NLL scores to quantify the anomaly uncertainty. 

\subsection{Quantifying anomaly uncertainties}

\begin{figure}[H]
\centerline{\includegraphics[scale=.55]{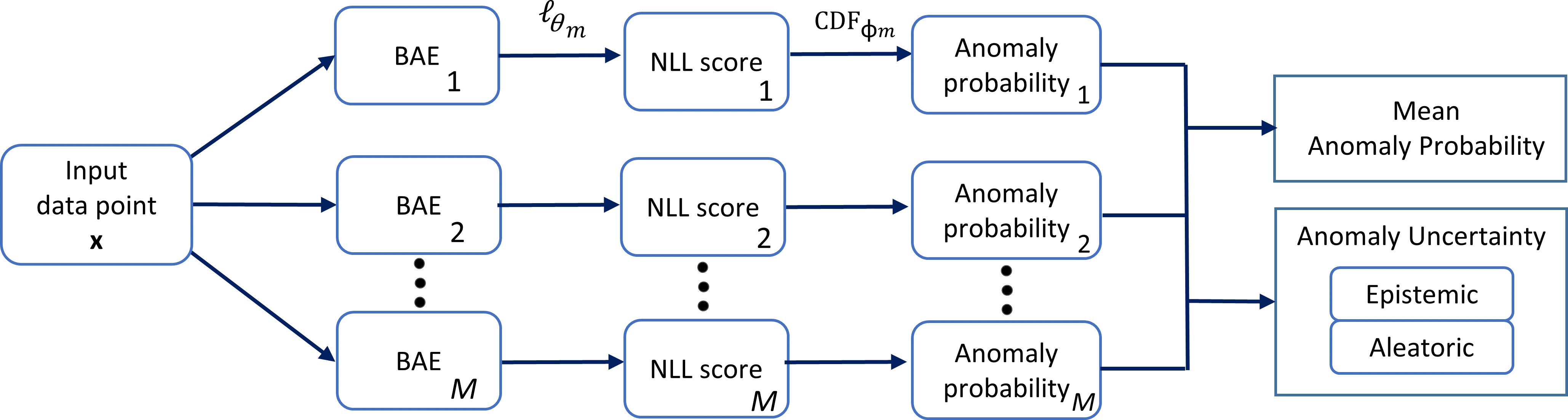}}
\caption{Workflow for quantifying the probability of an anomalous event and the predictive uncertainty using an ensemble of $M$-samples from the BAE posterior.}
\label{fig:proba-workflow}
\end{figure}

\textbf{Overview.} The proposed method for quantifying anomaly uncertainty is illustrated in \cref{fig:proba-workflow}. Although we use the NLL as an anomaly score, it is not a \textit{true} probability of an anomalous outcome. Hence, we first convert the NLL scores into anomaly probabilities via the cumulative distribution function (CDF). Next, using the anomaly probabilities, we compute the epistemic and aleatoric uncertainties. Capturing both uncertainties is crucial for forming a holistic quantification of predictive uncertainty; losing one of them leads to incomplete quantification and hence a lower quality of uncertainty estimate. Summing them forms the total anomaly uncertainty. Now, we formally describe our method for obtaining the total anomaly uncertainty with the BAE.

\textbf{(i) Distribution of anomaly scores.} 

Let $\phi_m$ be the vector of parameters of the distribution of NLL scores from the $m$-th BAE posterior sample,

\begin{equation}
    \NLLfunc{\textbf{x})} \thicksim \text{Q}(\phi_m)
\end{equation}
where Q is an arbitrary distribution. The optimal choice of distribution Q is dataset dependent, and in this study, we have experimented with setting Q as either a Gaussian, exponential or uniform distribution. For simplicity, we estimate $\phi_m$ by applying maximum likelihood estimation on the anomaly scores computed over the training data, i.e. $\hat{\phi}_m = \operatorname*{arg\,max}_{\phi_m}\prod_{i=1}^Np(\NLLfunc{\textbf{x}_i^{train})}\,|\,{\phi_m})$.

\textbf{(ii) Quantification of anomaly probability.} We define $y \in \{0,1\}$ as the outcome of an anomaly for a new data point $\textbf{x}^*$, which has a Bernoulli distribution,
\begin{equation} \label{eq:ber-y-anomaly}
    y | \textbf{x*}, X^{train}, \hat{\theta}_m, \hat{\phi}_m \thicksim \text{Ber}(\panomaly)
\end{equation}
where the anomaly probability $\panomaly$ is modelled using the CDF,
\begin{equation}
    \panomaly = \text{CDF}_{\hat{\phi}_m}(\NLLfunc{\textbf{x}^*)})
\end{equation}
Using the CDF to quantify anomaly probability is not uncommon; applying a min-max scaler \cite{scikit-learn} on the anomaly scores to rescale into $[0,1]$ is the same as fitting to a uniform distribution and using its CDF.

Alternatively, the empirical cumulative distribution function (ECDF) can be used to estimate $\panomaly$. The ECDF essentially captures the fraction of anomaly scores of the training data with lower values than the observed test anomaly score. Moreover, the ECDF converges to the true CDF as the number of examples increases according to Glivenko-Cantelli theorem \cite{tucker1959generalization}. As such, the ECDF benefits from an increasing number of training data, and we can avoid having to specify a distribution Q for the anomaly scores by leveraging the ECDF.
\begin{equation}
    \text{ECDF}(\NLLfunc{\textbf{x}^*)}) = \frac{1}{N}\sum_{i=1}^N1_{\NLLfunc{\textbf{x}_i^{train})}\,\leq\,{\NLLfunc{\textbf{x}^*)}}}
\end{equation}
In addition to using the CDF or ECDF when converting anomaly score to probability, we may choose to apply a customised scaling method proposed by Kriegel et al. \cite{kriegel2011interpreting},
\begin{equation}
    p_m^{Z_{anomaly}}(\textbf{x}^*) = \text{max}\bigg\{0,\frac{\text{CDF}_{\hat{\phi}_m}(\NLLfunc{\textbf{x}^*)})-\text{CDF}_{\hat{\phi}_m}(\E[\NLLfunc{{\textbf{x}^{train}})}])}{1-\text{CDF}_{\hat{\phi}_m}(\E[\NLLfunc{{\textbf{x}^{train}})}])}\bigg\}
\end{equation}
reflecting the beliefs that the anomaly probability is zero when the observed anomaly score is less than or equal to the average anomaly score of the training set (i.e. $p_m^{Z_{anomaly}}(\textbf{x}^*) = 0$ for $\text{CDF}_{\hat{\phi}_m}(\NLLfunc{\textbf{x}^*)}) \leq{\text{CDF}_{\hat{\phi}_m}(\E[\NLLfunc{{\textbf{x}^{train}})}])}$), and that the anomaly probability increases as the anomaly score gets higher than the average anomaly score. 

\begin{figure}[H]
\centerline{\includegraphics[width=.50\textwidth]{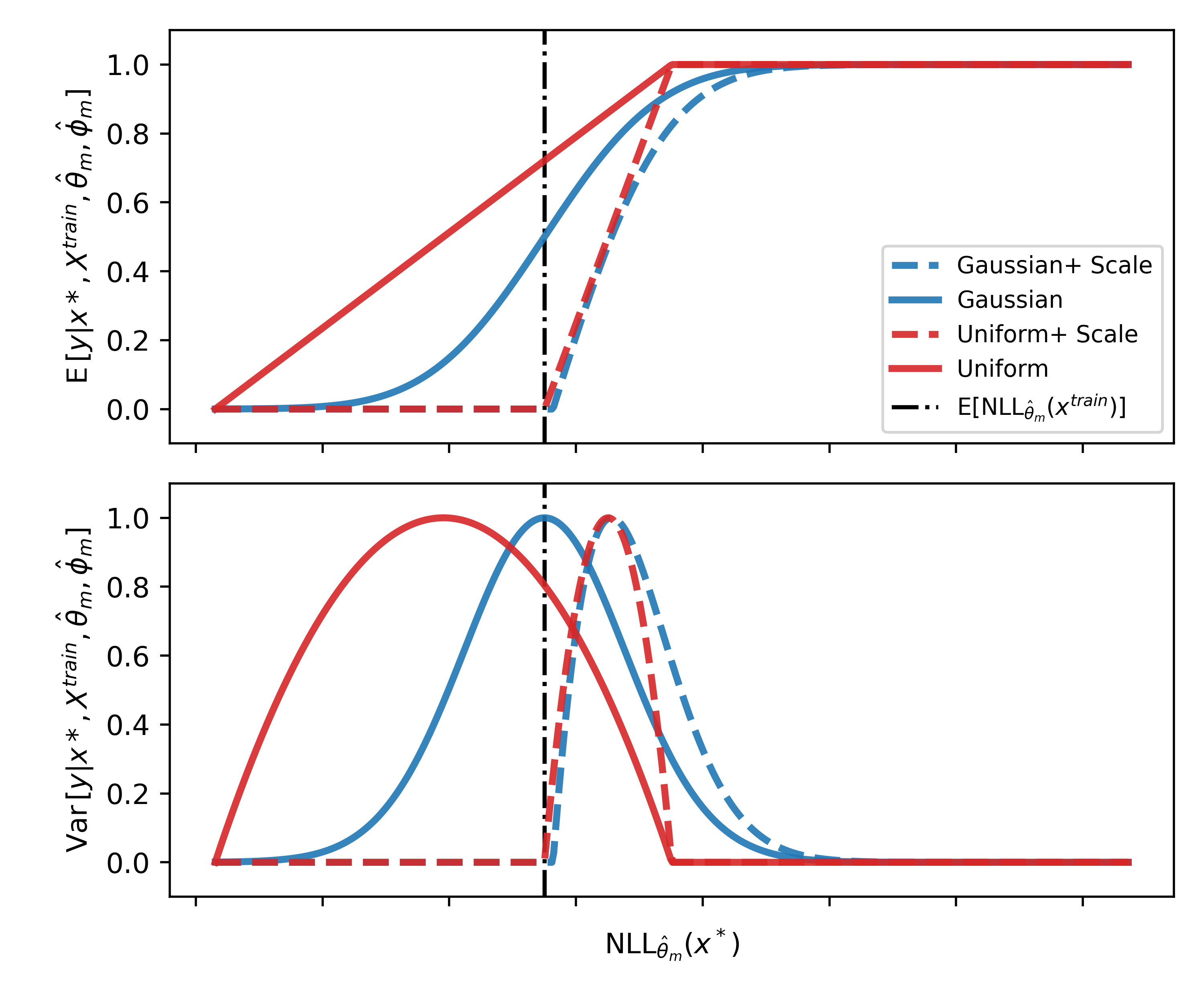}}
\caption{Conversion of $\NLLfunc{\textbf{x}^*)}$ from a single sample of BAE into $\E[y|\textbf{x*}, X^{train}, \hat{\theta}_m, \hat{\phi}_m]$ (upper panel) and $\text{Var}[y|\textbf{x*}, X^{train}, \hat{\theta}_m, \hat{\phi}_m]$ (lower panel). Distribution Q is chosen to be either the Gaussian or uniform distribution with an option of applying the customised scaling method. The customised scaling shifts the graph to the right by treating $\E[\NLLfunc{{\textbf{x}^{train}})}]$ as a reference.}
\label{fig:visual-proba-conv}
\end{figure}

By definition, the mean and variance of the Bernoulli distribution in \cref{eq:ber-y-anomaly} are:
\begin{gather}
\E[y|\textbf{x*}, X^{train}, \hat{\theta}_m, \hat{\phi}_m] = \panomaly \label{eq:ex-ber}\\ 
\text{Var}[y|\textbf{x*}, X^{train}, \hat{\theta}_m, \hat{\phi}_m] = \panomaly(1-\panomaly)  \label{eq:var-ber}
\end{gather}
See \cref{fig:visual-proba-conv} for an illustrative example of \cref{eq:ex-ber} and \cref{eq:var-ber}.

Next, to predict an anomalous event, we summarise the anomaly probabilities from all BAE posterior samples via expectation,
\begin{equation}
    \E[y|\textbf{x*}, X^{train}] \approx{} \E{}_{{\theta},{\phi}}[\,\E[y|\textbf{x*}, X^{train}, \hat{\theta}_m, \hat{\phi}_m]\,]= \frac{1}{M}\sum_{m=1}^{M}\panomaly
\end{equation}
A reasonable threshold can be set for the mean anomaly probability at any point where $\E[y|\textbf{x*}, X^{train}] \ge 0.5$ to get a hard prediction of an anomalous event. Moreover, as a calibrated anomaly score, its value ranges in [0,1] and facilitates better interpretation of an anomalous event than the raw anomaly score, which takes on any real number \cite{kriegel2011interpreting}. 

\textbf{(iii) Decomposition of anomaly uncertainty.} Based on the law of total variance \cite{weiss2005course}, we decompose the total anomaly uncertainty, \utotal into its epistemic and aleatoric components, 
\begin{equation} \label{eq:utotal} 
    \begin{split}
        \text{Var}[y|\textbf{x*}, X^{train}] \approx{ \underbrace{\E{}_{{\theta},{\phi}}[\text{Var}[y|\textbf{x*}, X^{train}, \hat{\theta}_m, \hat{\phi}_m]]}_{\text{\ualea}}}\,+ \\ \underbrace{\text{Var}_{{\theta},{\phi}}[\E{}[y|\textbf{x*}, X^{train}, \hat{\theta}_m, \hat{\phi}_m]]}_{\text{\uepi}}
    \end{split}
\end{equation}
Substituting with \cref{eq:ex-ber} and \cref{eq:var-ber}, the \ualea and \uepi can be computed as
\begin{gather} 
\E{}_{{\theta},{\phi}}[\text{Var}[y|\textbf{x*}, X^{train}, \hat{\theta}_m, \hat{\phi}_m]] = 
\E{}_{m}[\{\panomaly(1-\panomaly)\}_{m=1}^M] \\
\text{Var}_{{\theta},{\phi}}[\E{}[y|\textbf{x*}, X^{train}, \hat{\theta}_m, \hat{\phi}_m]] = 
\text{Var}_{m}[\{\panomaly\}_{m=1}^M]
\end{gather}
Our method propagates the model uncertainty from $M$ samples of the BAE posterior to yield a holistic quantification of anomaly uncertainty. In contrast, when we have a deterministic AE (i.e. $M=1$), the \uepi term is zero and the remaining component is the \ualea, signifying the inability of the deterministic AE to capture epistemic uncertainty.

\textbf{Note on interpretability.} Reading the \utotal can be unintuitive since the variance of a Bernoulli distribution ranges in [0,0.25]. Therefore, to enhance interpretability, we multiply \utotal $\times 4$ to range between [0,1] in this study. 
\subsection{Alternative ways to quantify anomaly uncertainties} \label{section:alt-unc}

We investigate two other methods for quantifying anomaly uncertainty. Firstly, by applying the method proposed by Perini et al. \cite{perini2020quantifying} on the BAE under the assumption that the training set is not polluted with anomalies, we arrive at the following uncertainty estimate,

\begin{equation} \label{eq:exceed}
    U^{exceed} = 
      \begin{cases} 
     (\E[y|\textbf{x*}, X^{train}])^N &,\, \E{}_{\theta}[\NLLfunc{\textbf{x}^*)}] < \text{max}\{\E{}_{\theta}[\NLLfunc{\textbf{x}^{train}_n)}]\}_{n=1}^{N} \\
      1-(\E[y|\textbf{x*}, X^{train}])^N &,\, \text{otherwise}
  \end{cases}
\end{equation}
where $N$ is the number of training examples. Like our method, the \uexceed relies on quantifying the $\panomaly$. The difference, however, is in the conversion from anomaly probability to anomaly uncertainty, which we posit a problem with the \uexceed is in the regime of high $N$, as this leads to extreme values of uncertainty estimate. In effect, it is prone to over- and underconfidence when the training examples are abundant, limiting its scalability to larger datasets. In addition, the \uexceed does not account for epistemic uncertainty, leading to a poorer uncertainty estimate.

\begin{figure}[H]
\centerline{\includegraphics[width=\textwidth]{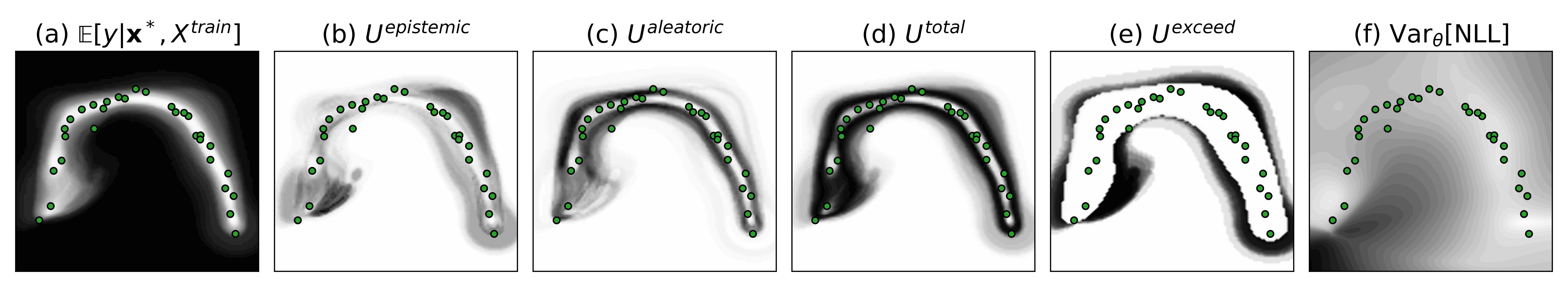}}
\caption{Visualisation of (a) $\E[y|\textbf{x*}, X^{train}]$ as anomaly scores, and measures of uncertainties using (b) \uepi, (c) \ualea, (d) \utotal, (e) \uexceed, and (f) \VarNLL using a BAE-Ensemble on a toy dataset with 2 features. Green dots are training examples while darker regions indicate higher values. \uepi and \ualea complement each other, forming reasonable regions of uncertainty with the \utotal. By contrast, \uexceed is overconfident in regions surrounding the training examples.}
\label{fig:visual-2d-unc}
\end{figure}

The second alternative way to measure anomaly uncertainty is by taking the variance across the $M$ samples of NLL scores from the BAE or VAE, 
\begin{equation}
    \text{Var}_{\theta}[\NLLfunc{\textbf{x}^*)}] = \frac{1}{M} \sum_{m=1}^{M}(\NLLfunc{\textbf{x}^*)}-\E{}_{\theta}[\NLLfunc{\textbf{x}^*)}])^2
\end{equation}
We shall refer to this method as \VarNLL. Note that \VarNLL is not applicable with the deterministic AE since the number of sampled predictions is 1 (\VarNLL=0). See \cref{fig:visual-2d-unc} for an illustration of the various methods for quantifying anomaly uncertainty. 

\subsection{Evaluating predictive uncertainty in anomaly detection}
 
We evaluate the quality of uncertainty in the task of classifying anomalies with a reject option. Specifically, in addition to classifying a data point as either inlier or anomaly, the BAE has the option of withholding its prediction by not assigning any label. The rationale is that predictions with high uncertainty are likely to be erroneous; hence rather than accepting all predictions in a high-stake application, it is safer to accept only predictions of low uncertainty, which are more accurate \cite{gao2019towards}.

In this setup, we use the accuracy-rejection curve (ARC) proposed by Nadeem et al. \cite{nadeem2009accuracy}, which plots the accuracy of the retained predictions against the rejection rate evaluated at multiple anomaly uncertainty thresholds. In effect, the ARC illustrates the trade-off between accuracy and rejection rate. As a measure of accuracy, while any metrics for binary classification can be used, we prefer the geometric mean of the sensitivity and specificity (\GSS) for evaluating classifiers on an imbalanced dataset \cite{kuncheva2019instance}. Then, to summarise the model's performance, we compute the weighted average accuracy 
\begin{equation}
    W^{GSS} = \frac{\sum_i^{T}(100-r_{i})GSS_{i}}{\sum_i^{T}(100-r_{i})}
\end{equation} 
where $T$ is the number of uncertainty threshold settings and $r$ is the rejection rate in percentage. For the anomaly uncertainty to be skillful, the \WGSS should be higher than the baseline \GSS, \baseGSS,  where $r=0$,  (i.e. not using the anomaly uncertainty). Hence, an important metric is the difference between the \WGSS and the baseline,
\begin{equation}
    Gain^{GSS} = W^{GSS}-Base^{GSS}
\end{equation}
which measures the average improvement in accuracy when uncertainty is included. It is possible for the \gainGSS to be negative in value, indicating a drop in performance due to the poor utility of uncertainty in rejecting erroneous predictions. 

\section{Experiment setup}

The full code for reproducing the results will be released upon acceptance of this paper.

\subsection{Datasets}
We use a collection of benchmark datasets, Outlier Detection Datasets (ODDS) \cite{rayana2016}, which are common in extant studies \cite{campos2016evaluation, chen2017outlier}. In addition, we use two publicly available datasets for industrial applications: ZeMA \cite{tizian2018} for condition monitoring and STRATH \cite{tachtatzis2019} for quality inspection. In the ZeMA dataset, there are 4 tasks; each task uses sensor measurements as inputs for detecting anomalies in hydraulic subsystems, namely, (i) the cooler, (ii) the valve, (iii) the pump, and (iv) the accumulator. While, in the STRATH dataset, the tasks are to detect defective parts manufactured from a radial forging facility. Different sensors are used for each task: (i) a position sensor (L-ACTpos), (ii) a speed sensor (A-ACTspd), (iii) a servo (Feedback-SPA), and (iv) all sensors from tasks (i-iii). See \cite{helwig2015, luo2020uncertainty} for detailed descriptions of the facility setup and data collection of ZeMA and STRATH datasets.

For all datasets, we split the inliers into train-test sets with a ratio of 70:30 and include all anomalies in the test set. We use the min-max scaler \cite{scikit-learn} to transform the input features into [0,1]. The resulting dimensions of the datasets is tabulated in \cref{table:datasets}. Additional details of preprocessing are included in \cref{app:data-pproc}.

\begin{table}[H]
\caption{Number of examples in train and test sets, and number of features for each task in ODDS, ZeMA and STRATH datasets.}
\centering
\resizebox{.8\textwidth}{!}{
    \begin{tabular}{@{}lcccc@{}}
    \toprule
    Datasets \& tasks & Train (inliers) & Test (inliers) & Test (anomalies) & Features \\ \midrule
    \textbf{ODDS} &  &  &  &  \\
    (i) Cardio & 1324 & 331 & 176 & 21 \\
    (ii) Lympho & 113 & 29 & 6 & 18 \\
    (iii) Optdigits & 4052 & 1014 & 150 & 64 \\
    (iv) Pendigits & 5371 & 1343 & 156 & 16 \\
    (v) Thyroid & 2943 & 736 & 93 & 6 \\
    (vi) Ionosphere & 180 & 45 & 126 & 33 \\
    (vii) Pima & 400 & 100 & 268 & 8 \\
    (viii) Vowels & 1124 & 282 & 50 & 12 \\ \midrule
    \textbf{ZeMA} &  &  &  &  \\
    (i) Cooler & 518 & 223 & 242 & 60 \\
    (ii) Valve & 787 & 338 & 30 & 60 \\
    (iii) Pump & 854 & 367 & 22 & 60 \\
    (iv) Accumulator & 419 & 180 & 272 & 60 \\ \midrule
    \textbf{STRATH} &  &  &  &  \\
    (i) L-ACTpos & 51 & 23 & 7 & 562 \\
    (ii) A-ACTspd & 51 & 23 & 7 & 562 \\
    (iii) Feedback-SPA & 51 & 23 & 7 & 562 \\
    (iv) All sensors & 51 & 23 & 7 & 1686 \\ \bottomrule
    \end{tabular}
    }
    \label{table:datasets}
\end{table}

\subsection{Model hyperparameters}

We fit the deterministic AE, VAE, and BAE models to the train set using the Adam optimiser  \cite{kingma2014adam} for 100 epochs with a fixed learning rate of 0.001 for STRATH. For ODDS and ZeMA, we do not use a fixed learning rate but instead employ an automatic learning rate finder \cite{smith2017cyclical}. The number of posterior samples is set to $M=100$ for the VAE, BAE-MCD and BAE-BBB, while $M=10$ for the BAE-Ensemble. The prior scaling term (weight decay) for all models is $1\times{10^{-10}}$. We specify the architecture of the encoder in \cref{table:architecture}. U-Net skip connections \cite{unet2015} are implemented in all models except for those in STRATH tasks (i) and (ii).

\begin{table}[ht] 
\centering
\caption{Encoder architecture for ODDS, ZeMA and STRATH datasets. We set the latent dimensions to be half of the flattened input dimensions. The decoder is a reflection of the encoder, in which the Conv1D layers are replaced by Conv1D-Transpose layers. We apply layer normalisation after each Conv1D and linear layer for ODDS and STRATH. The leaky ReLu \cite{maas2013rectifier} is used as the activation function with a slope of 0.01 while the sigmoid function is used at the decoder's final layer.}
    \begin{subtable}[h]{\textwidth}
        \centering
        \scriptsize \caption{ODDS}
            \resizebox{0.6\textwidth}{!}{
                    \begin{tabular}{@{}cccc@{}}
                    \toprule
                    Layer       & Output channels/nodes & Kernel & Strides \\ \midrule
                    Linear       & Input dimensions $\times 4$      & -      & -      \\ 
                    Linear       & Input dimensions $\times 4$       & -      & -      \\ 
                    Linear       & Latent dimensions       & -      & -      \\ \bottomrule
                    \end{tabular}
            }
    \end{subtable}
    \hfill
    \begin{subtable}[h]{0.6\textwidth}
        \centering
        \vskip 0.2in
        \scriptsize \caption{ZeMA and STRATH}
            \resizebox{\textwidth}{!}{
                \begin{tabular}{@{}cccc@{}}
                \toprule
                Layer       & Output channels/nodes & Kernel & Strides \\ \midrule
                Conv1D & 10      & 8  & 2   \\
                Conv1D & 20      & 2  & 2   \\
                Reshape     & -              & -      & -       \\
                Linear       & 1000       & -      & -      \\ 
                Linear       & Latent dimensions       & -      & -      \\ \bottomrule
                \end{tabular}
            }
    \end{subtable}
     \label{table:architecture}
\end{table}

We repeat the experiment 10 times with different random seeds. For each run, we vary the choice of anomaly score distribution Q to be either the Gaussian, exponential, or uniform distribution. Due to the large combinations of methods, we report results where Q is implicitly chosen (unless stated) to maximise the \WGSS criterion.

\section{Results and discussion} 

\begin{figure}[H]
	\centering
	\begin{subfigure}[t]{\textwidth}
	\caption{ODDS}\label{fig:misclas-a}
    \includegraphics[width=\textwidth]{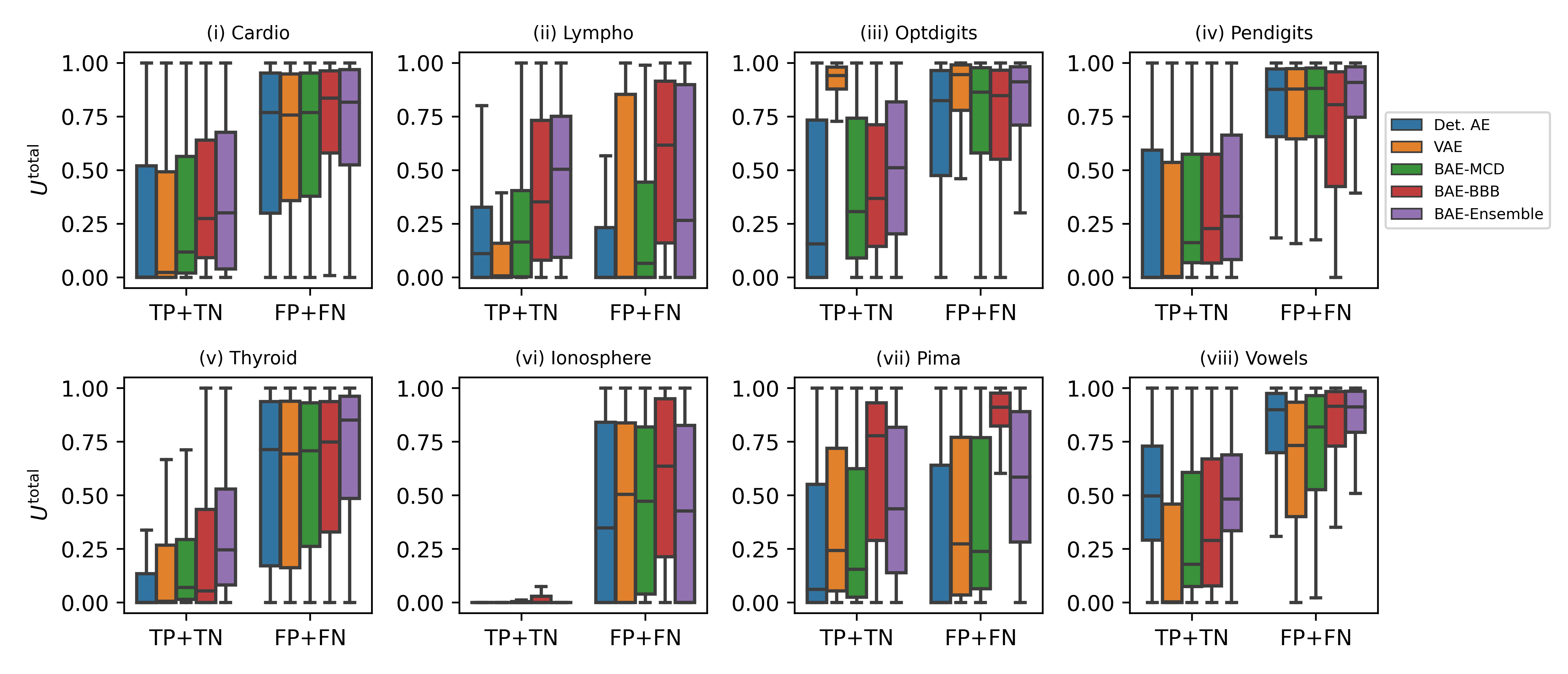}
	\end{subfigure}
	\quad
	\begin{subfigure}[t]{\textwidth}
	\caption{ZeMA}\label{fig:misclas-b}
    \includegraphics[width=\textwidth]{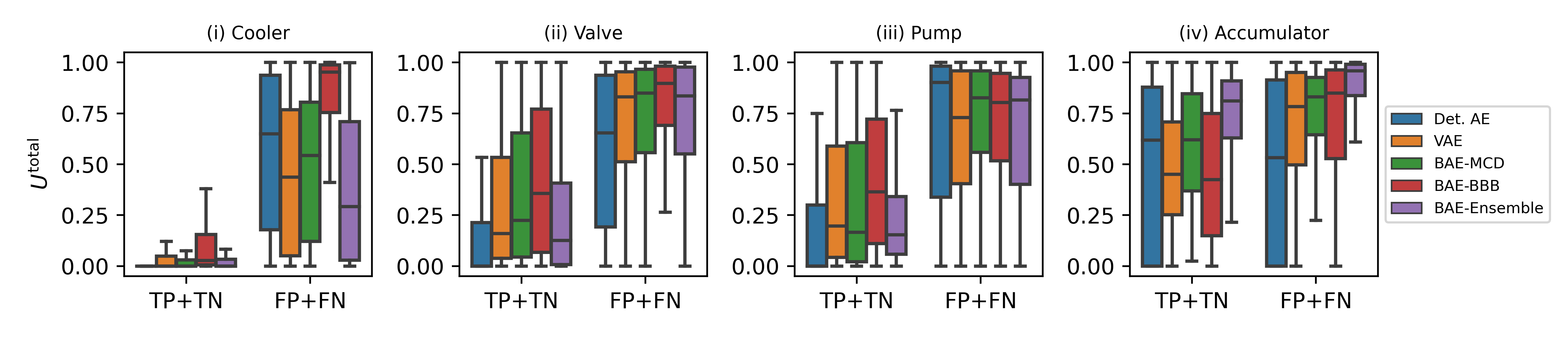}
	\end{subfigure}
	\quad
	\begin{subfigure}[t]{\textwidth}
	\caption{STRATH}\label{fig:misclas-c}
    \includegraphics[width=\textwidth]{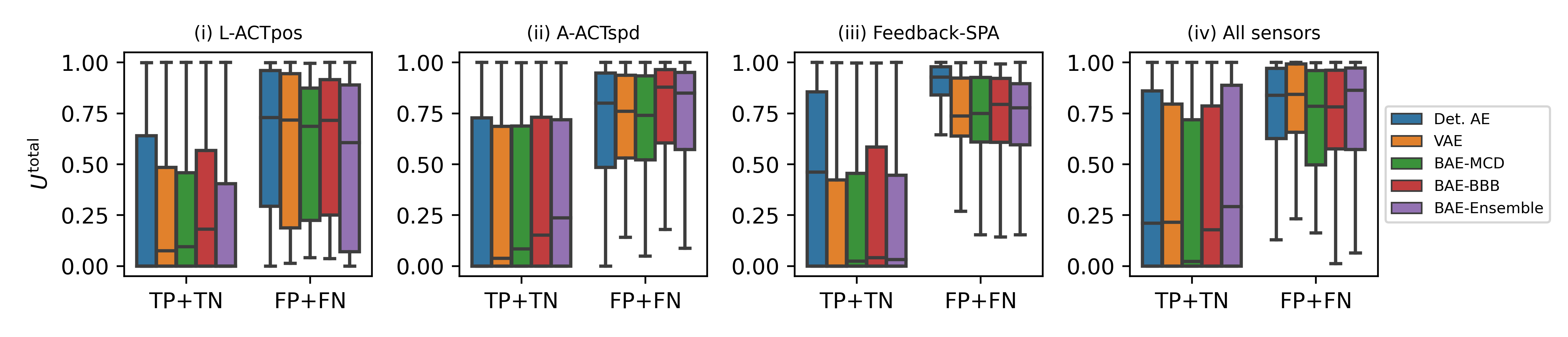}
	\end{subfigure}
	\caption{Box plots of \utotal for correct predictions (TP+TN) and erroneous predictions (FP+FN) using various models. (TP = true positives, TN = true negatives, FP = true positives, FN = true negatives).}\label{fig:misclas}
\end{figure}

\textbf{Incorrect predictions have higher uncertainty.} From \cref{fig:misclas}, we observe that inaccurate predictions are usually accompanied by higher \utotal scores, demonstrating the use of \utotal to anticipate both Type I and Type II predictive errors \cite{banerjee2009hypothesis}, in the absence of true labels. Nonetheless, this is not always the case; for example, in \cref{fig:misclas-a}(iii) the VAE assigns similar or higher \utotal scores to correct predictions. Likewise, the deterministic AE does the same for ZeMA task (iv). In such cases, the uncertainty quality is poor, highlighting the ongoing research challenge of obtaining reliable uncertainty estimates. Hence, a careful assessment of uncertainty quality is required before deploying in production.      

\begin{figure}[H]
	\centering
		\begin{subfigure}[H]{\textwidth}
	\caption{ODDS}\label{fig:arc-models-a}
    \includegraphics[width=\textwidth]{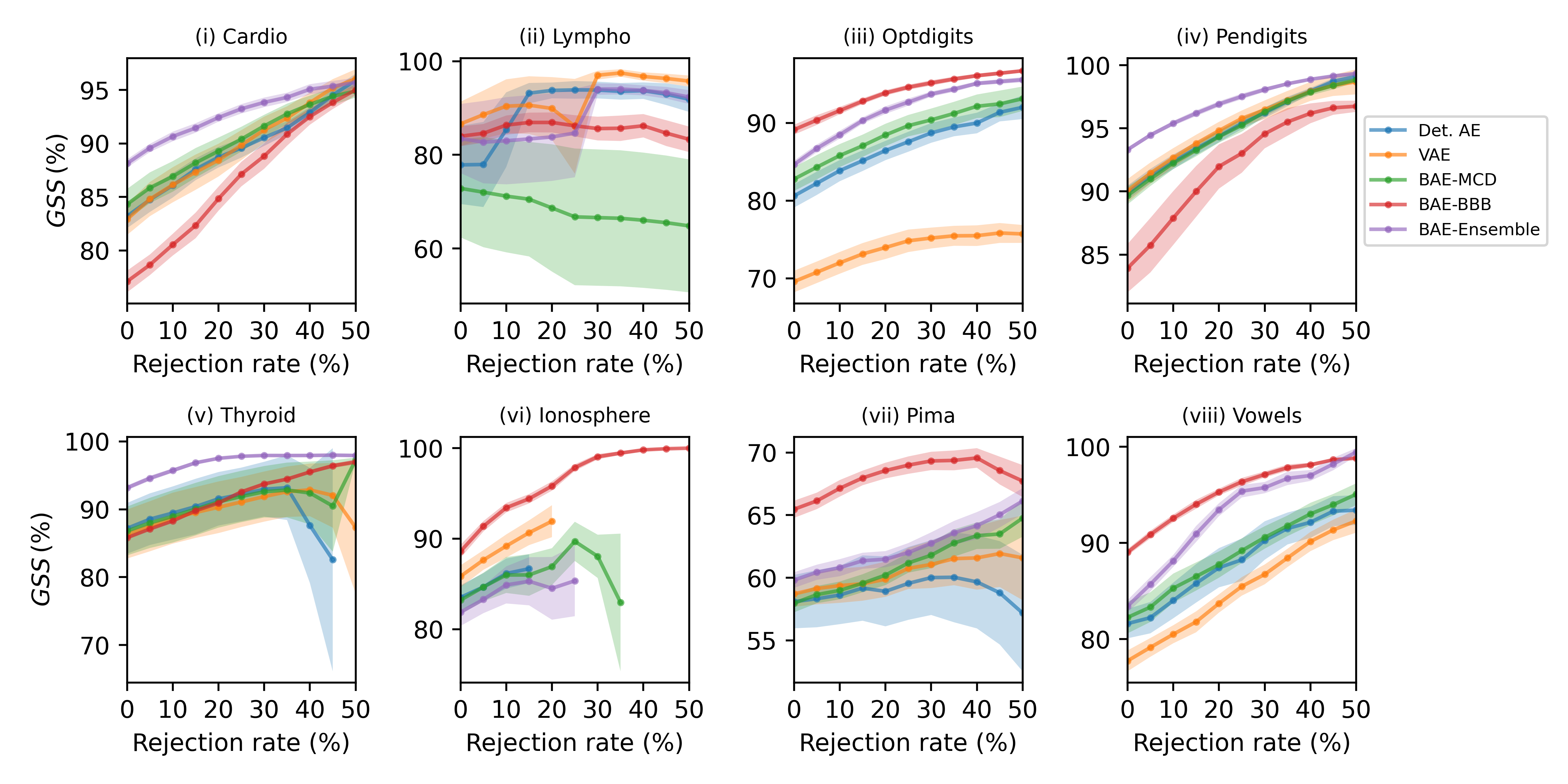}
	\end{subfigure}
	\quad
	\begin{subfigure}[H]{\textwidth}
	\caption{ZeMA}\label{fig:arc-models-b}
    \includegraphics[width=\textwidth]{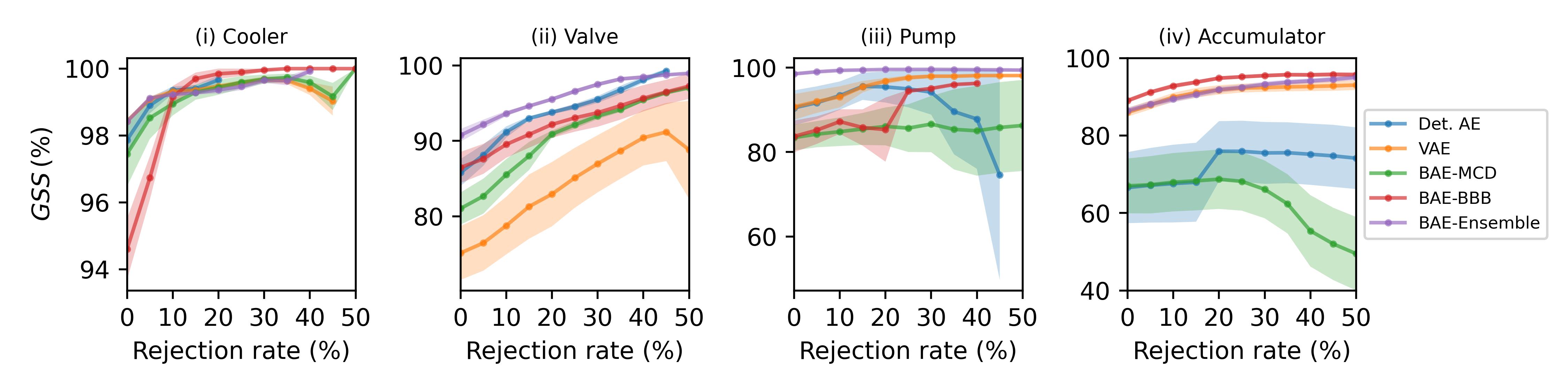}
	\end{subfigure}
	\quad
	\begin{subfigure}[H]{\textwidth}
	\caption{STRATH}\label{fig:arc-models-c}
    \includegraphics[width=\textwidth]{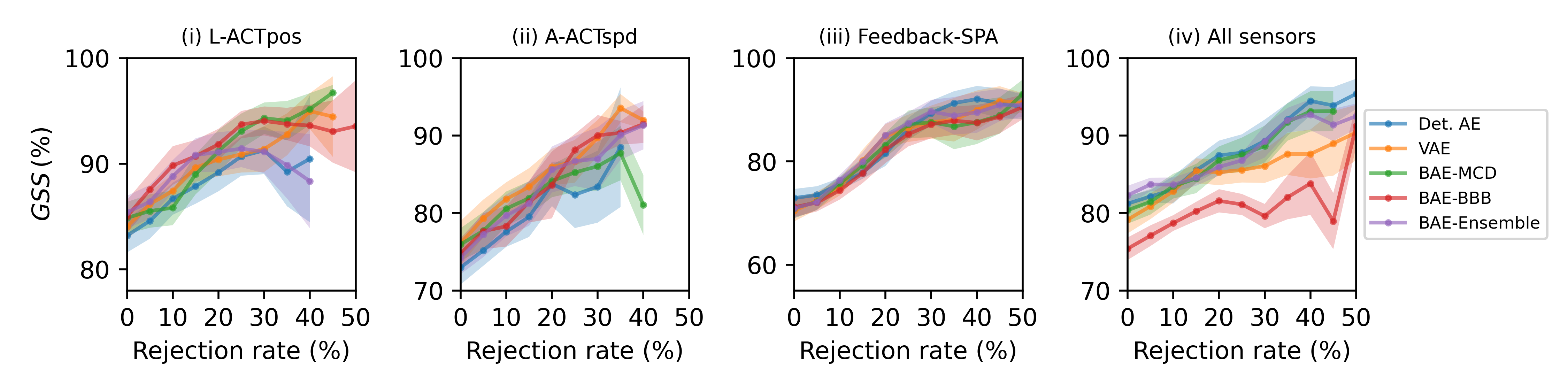}
	\end{subfigure}
	\caption{ARCs for comparisons of deterministic AE, VAE and BAEs with posterior approximated under various techniques. Mean and standard error of GSS are evaluated on (a) ODDS, (b) ZeMA and (c) STRATH datasets over 10 experiment runs. \utotal is used as the rejection criterion with the Q distribution chosen based on the best \WGSS score. }\label{fig:arc-models}
\end{figure}

\textbf{Rejecting uncertain predictions may yield higher predictive accuracy.}  Strikingly, the detection performance can rise to \textgreater{\,90\%} \GSS after rejecting 40\% of uncertain predictions for most tasks (\cref{fig:arc-models}). Without rejection, however, for instance, in \cref{fig:arc-models-c}(iii), most models scored much lower \GSS $(\approx 70\%)$. Note that not all models yield positive performance gains as we reject predictions of high uncertainty, shown by a constant or deteriorating trend, e.g. BAE-MCD on \cref{fig:arc-models-a}(ii) and \cref{fig:arc-models-b}(iv)). 

\begin{figure}[htbp]
    \centerline{\includegraphics[scale=.70]{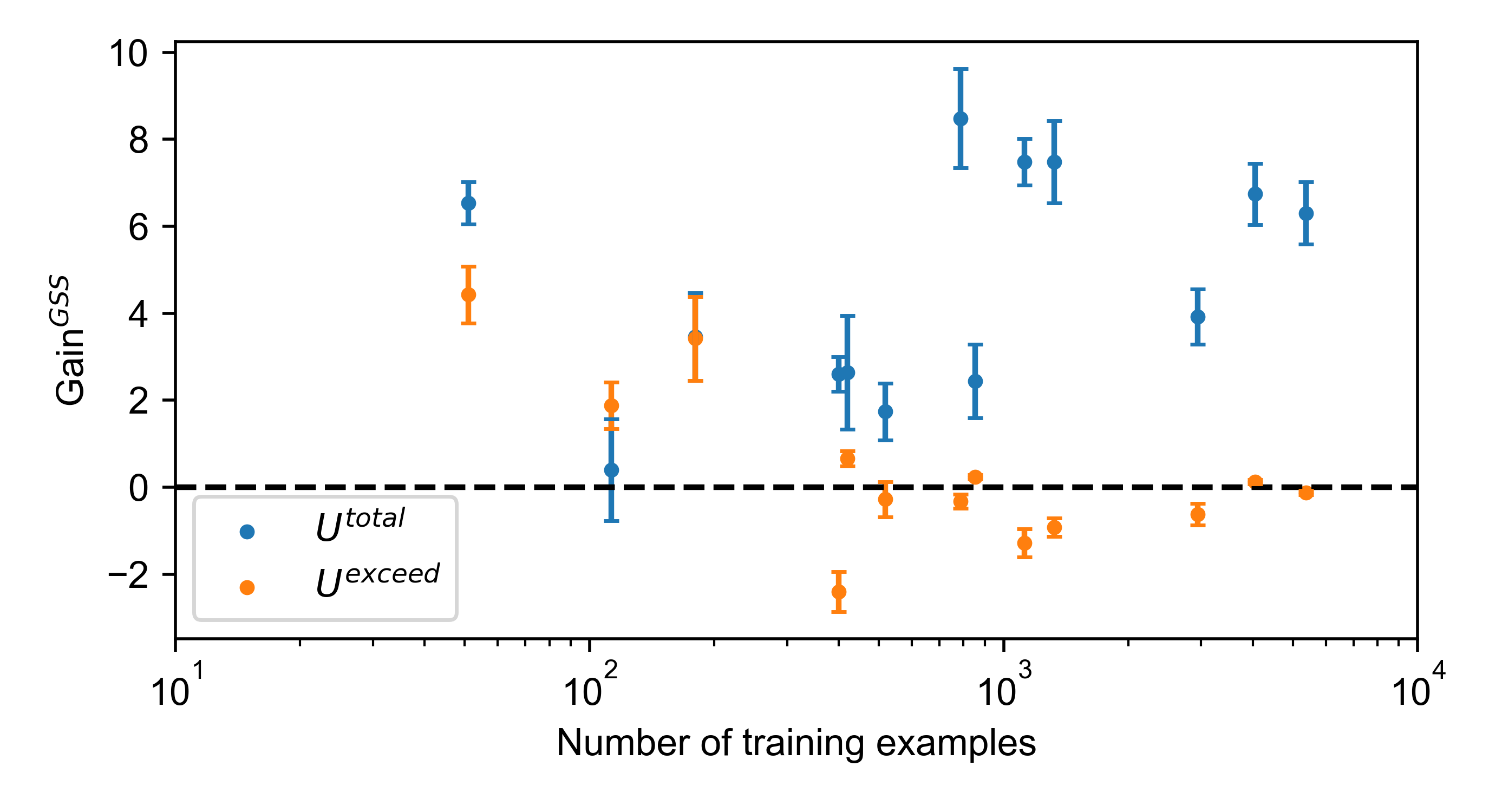}}
    \caption{Mean and standard error of $Gain^{GSS}$ against the number of training examples in ODDS, ZeMA and STRATH datasets. Statistics are calculated over 10 experiment runs and 5 types of AEs. Dotted line at $Gain^{GSS}=0$ is the baseline.} \label{fig:trainsize}
\end{figure}

\textbf{Should we use the \uexceed or \VarNLL method to quantify anomaly uncertainty?} Earlier, we mentioned that \uexceed (\cref{eq:exceed}) does not scale well with the number of training examples. We notice it has a poor mean score of 0.1\% $Gain^{GSS}$ on the ZeMA dataset (\cref{table:retained}) while having a better score on the STRATH dataset, which has fewer training examples than ZeMA. Further inspection confirms the performance of \uexceed is poor and does not scale well with number of training examples \cref{fig:trainsize}. This lack of scalability makes the \uexceed less practical when there is an abundance of training data, or when data is incrementally collected. On the other hand, we do not observe such behaviours with the \utotal as it performs well even as training examples increase in number.

All in all, \utotal outperforms \uexceed and \VarNLL. Despite its implementation simplicity, the \VarNLL is unreliable as a measure of anomaly uncertainty as we find that \VarNLL yields the lowest scores for both ZeMA and STRATH datasets in \cref{table:retained}. The \utotal also performs better than \uepi and \ualea, proving the benefit of combining both types of uncertainties compared to using either one of them.  

\begin{table}[H] 
\caption{Mean \WGSS ($Gain^{GSS}$) is evaluated evaluated on (a) ODDS, (b) ZeMA and (c) STRATH datasets. Each dataset consists of multiple tasks and 10 experiment runs. Uncertainty estimation method with the highest mean \WGSS score is bolded. Values are shown in percentage.}
    \begin{subtable}[h]{\textwidth}
        \centering
        \scriptsize \caption{ODDS}
            \resizebox{\textwidth}{!}{
                \begin{tabular}{@{}llllll@{}}
                \toprule
                Model & \uepi & \ualea & \utotal & \uexceed & \VarNLL \\ \midrule
Deterministic AE & - & 84.6(+4.3) & 84.6(+4.3) & 80.2(0.0) & - \\
VAE & 81.3(+1.8) & 83.8(+4.1) & 84.3(+4.6) & 79.4(-0.3) & 78.4(+0.1) \\
BAE-MCD & 81.4(+1.7) & 83.8(+3.8) & 84.2(+4.2) & 79.9(0.0) & 77.9(-1.3) \\
BAE-BBB & 86.7(+4.0) & 87.8(+4.9) & 89.2(+6.3) & 83.2(0.0) & 80.5(-0.7) \\
BAE-Ensemble & 85.2(+1.5) & 86.4(+3.0) & 88.0(+4.6) & 84.0(+0.3) & 81.2(-0.8) \\ \midrule
\textbf{Mean} & 83.6(+2.2) & 85.3(+4.0) & \textbf{86.1(+4.8)} & 81.3(0.0) & 79.5(-0.7) \\ \bottomrule
                \end{tabular}
            }
    \end{subtable}
    \hfill
    
    \begin{subtable}[h]{\textwidth}
        \centering
        \vskip 0.2in
        \scriptsize \caption{ZeMA}
            \resizebox{\textwidth}{!}{
                \begin{tabular}{@{}llllll@{}}
                \toprule
                Model & \uepi & \ualea & \utotal & \uexceed & \VarNLL \\ \midrule
Deterministic AE & - & 88.1(+2.9) & 88.1(+2.9) & 85.4(+0.4) & - \\
VAE & 89.4(+1.2) & 92.6(+5.0) & 92.9(+5.4) & 88.2(+0.2) & 89.4(+1.4) \\
BAE-MCD & 82.8(-0.1) & 86.1(+3.8) & 85.8(+3.6) & 83.2(+0.3) & 84.4(+1.8) \\
BAE-BBB & 89.8(+2.4) & 91.9(+3.5) & 92.6(+4.1) & 88.0(-0.4) & 86.5(-0.1) \\
BAE-Ensemble & 94.9(+1.3) & 95.1(+1.6) & 96.7(+3.2) & 93.5(0.0) & 94.0(+1.1) \\ \midrule
\textbf{Mean} & 89.2(+1.2) & 90.8(+3.4) & \textbf{91.2(+3.8)} & 87.7(+0.1) & 88.6(+1.0) \\ \bottomrule
                \end{tabular}
            }
    \end{subtable}
    \hfill

    \begin{subtable}[h]{\textwidth}
        \centering
        \vskip 0.2in
        \scriptsize \caption{STRATH}
            \resizebox{\textwidth}{!}{
                \begin{tabular}{@{}llllll@{}}
                \toprule
                Model & \uepi & \ualea & \utotal & \uexceed & \VarNLL \\ \midrule
Deterministic AE & - & 83.9(+6.3) & 83.9(+6.3) & 81.4(+2.5) & - \\
VAE & 82.6(+3.7) & 84.1(+6.2) & 84.2(+6.7) & 81.2(+4.1) & 81.0(+1.6) \\
BAE-MCD & 83.7(+5.7) & 84.2(+6.2) & 84.5(+6.5) & 80.2(+4.2) & 79.0(+0.3) \\
BAE-BBB & 82.6(+5.5) & 82.4(+5.8) & 83.2(+6.7) & 80.4(+6.4) & 78.3(+0.1) \\
BAE-Ensemble & 81.9(+2.5) & 84.9(+6.7) & 84.7(+6.5) & 81.8(+4.9) & 80.0(0.0) \\ \midrule
\textbf{Mean} & 82.7(+4.4) & 83.9(+6.2) & \textbf{84.1(+6.5)} & 81.0(+4.4) & 79.6(+0.5) \\ \bottomrule

                \end{tabular}
            }
       
     \end{subtable}
     
     \label{table:retained}
\end{table}

\textbf{Deterministic AE vs stochastic AEs (VAE and BAE).} We discuss differences between AE models when using the \utotal to estimate the predictive uncertainty. The deterministic AE, which quantifies only the aleatoric component, stands as a baseline against the stochastic AEs. For most models, the stochastic AEs have a higher \WGSS than the deterministic AE, showing the benefits of accounting for model uncertainty in a probabilistic framework. However, this improvement does not hold for some models; one reason could be the lower quality of posterior sampling \cite{yao2019quality, pearce2020uncertainty}. On the other hand, the BAE-Ensemble outperforms the baseline AE and the VAE, scoring the highest \WGSS on ZeMA and STRATH, and second highest \WGSS on ODDS. 

Besides the performance gain, the BAE-Ensemble is relatively easy to implement compared to other stochastic AEs. It does not require any changes to the architecture or the optimiser of a deterministic AE, allowing a seamless transition from a deterministic AE to a stochastic AE. While the BAE does require $M$ times more computations than the deterministic AE, the computations are highly parallelisable since we can compute each ensemble member independently. Hence, the BAE can leverage the development of dedicated hardware for NN computations, e.g. edge computing devices for industrial environments \cite{li2019edge}.

\textbf{Effect of anomaly probability conversion.} Quantifying \utotal depends on the method of anomaly probability conversion (e.g. using either the CDF or ECDF, and whether to apply customised scaling); the switch from one method of conversion to the other strongly influences the quality of uncertainty. Notably, we find that customised scaling does improve the quality of uncertainty (see \cref{table:dist-positives}) as we notice large gains in the percentage of positive experiment runs after applying the customised scaling. 

\begin{table}[H]
\centering
\caption{Percentage of experiment runs where \baseGSS \textgreater{0} when using \utotal while varying the method of anomaly probability conversion. Percentage greater than 80\% are in bold.}

\scriptsize 

\resizebox{.90\textwidth}{!}{%
\begin{tabular}{@{}lccc@{}}
\toprule

\multicolumn{1}{c}{Conversion method} & \multicolumn{3}{c}{ \% (Positives/Total runs)} \\ \cmidrule(l){2-4} 
\multicolumn{1}{c}{} & ODDS & ZeMA & STRATH \\
\midrule

ECDF&18.2\% (73/400)&41.5\% (83/200)&48.5\% (97/200)\\ 
ECDF + Scale&\textbf{83.0\% (332/400)}&\textbf{94.5\% (189/200)}&\textbf{89.5\% (179/200)}\\ \midrule
Gaussian&37.8\% (151/400)&47.5\% (95/200)&50.5\% (101/200)\\ 
Gaussian + Scale&\textbf{80.5\% (322/400)}&69.0\% (138/200)&\textbf{90.0\% (180/200)}\\ \midrule
Exponen.&16.2\% (65/400)&48.5\% (97/200)&41.0\% (82/200)\\ 
Exponen. + Scale&\textbf{82.5\% (330/400)}&\textbf{89.0\% (178/200)}&44.0\% (88/200)\\ \midrule
Uniform&45.8\% (183/400)&22.0\% (44/200)&50.5\% (101/200)\\ 
Uniform + Scale&50.0\% (200/400)&25.5\% (51/200)&61.5\% (123/200)\\ \bottomrule

\end{tabular}
}
\label{table:dist-positives}
\end{table}

In cases where \utotal did not perform as well, we suggest one reason is the improper choice of distribution of NLL scores for the dataset. For instance, although the exponential distribution worked reliably well for ODDS and ZeMA, it did not work well for the STRATH dataset. Instead of committing to a specific distribution, we can use the ECDF, a non-parametric method, which works reliably well for all datasets. We report additional results in \cref{app:res-uncood}.

\section{Conclusion}

In this work, we have formulated the BAEs to estimate the predictive uncertainty for anomaly detection. Our method computes the \uepi and \ualea, and adding the two uncertainties forms the \utotal. 

Our experiments have evaluated the quality of uncertainty in the regime where the model is given the option to reject predictions of high uncertainty, and have demonstrated improvements using the proposed methods on multiple publicly available datasets. The best performing BAE consistently outperform the deterministic AE, highlighting the benefit of the Bayesian formulation.

Several directions for future work are possible. There is a need to further understand the conditions in which the anomaly uncertainty may fail, for instance, in adversarial attacks. Comprehensive case studies are necessary to understand the practical challenges in deploying BAEs. Future work should also explore the use of anomaly uncertainty for incremental and active learning.

\section*{Acknowledgments}

The work reported here was supported by the European Metrology Programme for Innovation and Research (EMPIR) under the project Metrology for the Factory of the Future (MET4FOF), project number 17IND12 and part-sponsored by Research England’s Connecting Capability Fund award CCF18-7157: Promoting the Internet of Things via Collaboration between HEIs and Industry (Pitch-In).

\bibliography{bib-unc_ood, bib-xai, bib-xiang, bib-icml2020}
\newpage
\appendix

\section{BAE posterior sampling methods} \label{app:methods}

We describe the following methods to sample from the BAE: BBB \cite{blundell2015weight}, MCD \cite{gal2016dropout} and anchored ensembling \cite{pearce2020uncertainty}.

\subsection{Variational inference} 

BBB and MCD are examples of variational inference. The idea of variational inference is to approximate the posterior distribution over parameters by introducing a variational distribution. 
\begin{equation}
    \qFull \approx \pFull
\end{equation}
where $\psi$ is the vector of parameters of the variational distribution, $q$. The objective of training is thus, to minimise the Kullback-Leibner (KL) divergence \cite{kullback1951information}, which is a measure of similarity between two distributions: the variational distribution and the true posterior. This yields an optimised variational distribution to be sampled from during prediction.

\begin{equation}
    \psi^{\mathrm{opt}} = \operatorname*{arg\,min}_{\psi}\mathrm{KL}[\qFull \,||\, \pFull] 
\end{equation}
The KL divergence can be approximated with $M$ samples of AE parameters $\{\theta_{m}\}_{m=1}^M$ from the variational distribution $q$. Note that minimising the KL divergence is the same as maximising the log evidence lower bound (ELBO).

\begin{equation}
    \mathrm{KL}[\qFull \,||\, \pFull] \approx \sum_{m=1}^{M}-\log{p(X|\theta_{m})}-\lambda\,
    \mathrm{KL}[\qSub \,||\, p(\theta_{m})]
\end{equation}
where the weight decay, $\lambda$ scales the KL divergence of the prior, effectively controlling the strength of regularisation. Note that the first term of the sum is the log-likelihood of a sample from the variational distribution.

Now, we discuss popular models of the variational distribution. In BBB, we use a diagonal Gaussian distribution as the variational distribution, while the prior is a mixture of 2 diagonal Gaussian distributions, 
\begin{gather*}
    \psi = (\mu,\sigma^2) \\
    \epsilon \sim \mathrm{N}(0,1) \\
    \theta = \mu + \sigma \epsilon
\end{gather*}
where $\epsilon$ is the noise term introduced for implementing the reparameterisation trick \cite{blundell2015weight}, necessary for backpropagating the gradients during optimisation. The KL prior loss is 
\begin{equation}
    \begin{split}
    \mathrm{KL}[\qSub \,||\, p(\theta_m)]^{\mathrm{BBB}} =
    \pi(\frac{||\theta_{m}||^2}{2\tau_1^2}-\log{\tau_1})\,+
    (1-\pi)(\frac{||\theta_{m}||^2}{2\tau_2^2}-\log{\tau_2)}\,+\\
        \frac{||\theta_{m}-\mu||^2}{2\sigma^2}-\log{\sigma}
    \end{split}
\end{equation}
where $\pi$ , $\tau_1$ and $\tau_2$ are parameters of the prior, which we fix in our experiments as 0.5, 1.0 and 0.1, respectively.

On the other hand, the MCD, which is also type of variational inference, is implemented by adding a dropout layer after each linear or convolutional layer. The corresponding variational distribution is a mixture of Gaussian and Bernoulli distributions.
\begin{gather*}
    \psi = W \\
    \epsilon \sim \mathrm{Ber}(1-p^\mathrm{dropout}) \\
    \theta = W \cdot \epsilon
\end{gather*}
where $p^\mathrm{dropout}$ is the dropout probability and $W$ refers to the AE parameters when they are not dropped out.  Further, the $p^\mathrm{dropout}$ is a hyperparameter and is not updated during training. We set $p^\mathrm{dropout}=0.01$ in our experiments. Lastly, the KL prior term can be approximated as

\begin{equation}
    \mathrm{KL}[\qSub \,||\, p(\theta_m)]^{\mathrm{MCD}} \approx
    ||\theta_{m}||^2
\end{equation}

\subsection{Anchored ensembling}

In anchored ensembling, posteriors are approximated by Bayesian inference under the family of methods called randomised maximum a posteriori (MAP) sampling, where model parameters are regularised by values drawn from a distribution (so-called anchor distribution), which can be set equal to the prior.

Assume our ensemble consists of $M$ independent autoencoders and each j-th autoencoder contains a set of parameters, $\theta_m$ where $m\in{\{1,2,3...M\}}$. In anchored ensembling for approximating the posterior distribution, the `anchored weights` for each autoencoder are unique and sampled during initialisation from a prior distribution $\theta_{m}^{\mathrm{anc}} \sim N(\mu_{m}^{\mathrm{anc}},\sigma_{m}^{\mathrm{anc}^2})$ and remain fixed throughout the training procedure. 

The autoencoders are trained by minimising the loss function, which is the negative sum of log-likelihood (based on i.i.d assumption) and log-prior where both are assumed to be Gaussian. For each member of the ensemble, the loss to be optimised is

\begin{equation} \label{eq_prior_loss}
\mathcal{L}(\theta_m,X) = -\log{p(X|\theta_m)} +\lambda||\theta_{m}-\theta_{m}^{\mathrm{anc}}||^2
\end{equation}
where $\lambda$ is a hyperparameter for scaling the regulariser term arising from the prior.

\section{Data preprocessing} \label{app:data-pproc}

\textbf{ODDS.} For each dataset in the ODDS collection, we do not apply any preprocessing steps aside from min-max scaling and data splitting. 

For ZeMA and STRATH, the inputs to the AE have the dimensions of $(B \times L \times K\,)$, where $B$ is the batch size, $L$ is the sequence length, and $K$ is the number of sensors. 

\textbf{ZeMA.} For each datapoint in the batch, The labels of each hydraulic subsystem's condition are available in the dataset, and we label the best working conditions as inliers while the remaining states as anomalies. We have chosen the following sensor-subsystem pairs in our analysis: temperature sensor (TS4) for the cooler, valve and accumulator, and pressure sensor (PS6) for the pump. We downsample the pressure sensor to 1Hz; no resampling is applied on the temperature data, hence $L=60$ and $K=1$. 

\begin{figure}[H]
\centerline{\includegraphics[scale=.425]{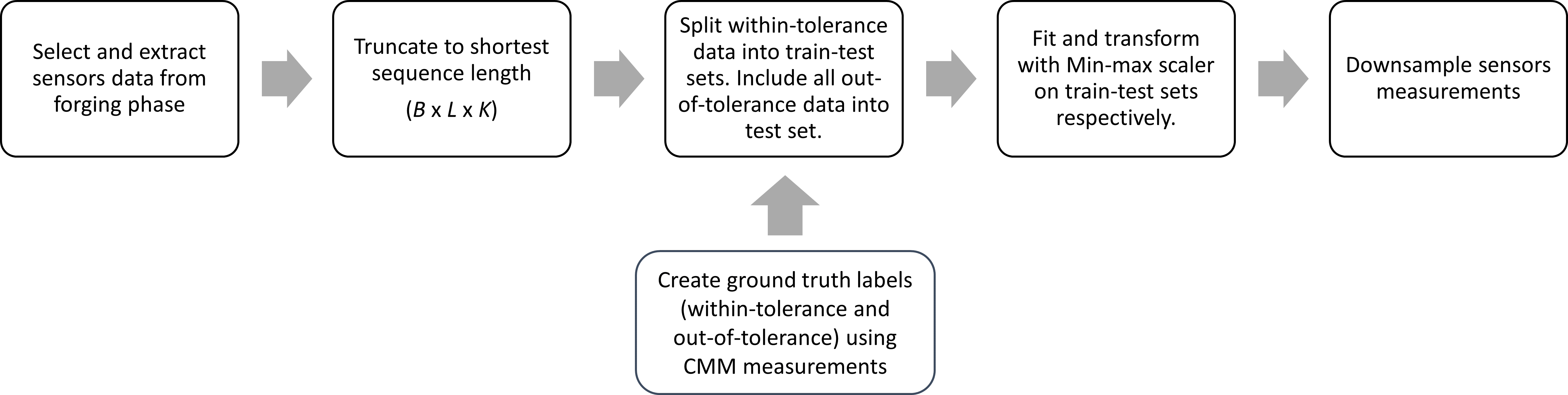}}
\caption{Data preprocessing pipeline for STRATH data set}
\label{fig:strath-preprocess}
\end{figure}

\textbf{STRATH.} The radial forging process is divided into the heating and forging phases. We segment the sensors data to consider only the forging phase. We downsample the data by 10 fold, reducing the sequence length to $L=562$ for all tasks. For tasks (i-iii), $K=1$ and $K=3$ for task (iv). For each forged part, measurements of its geometric dimensions are available as quality indicators; we focus our analysis on the \textit{38 diameter@200} dimension. To flag parts as anomalies, we first obtain the absolute difference between the measured and nominal dimensions, and subsequently, we apply the Tukey's fences method \cite{tukey1977exploratory}. The remaining parts are labelled as inliers. The data preprocessing steps are outlined in \cref{fig:strath-preprocess}. 

As mentioned, we apply min-max scaling for all datasets. For ODDS, we apply the min-max scaling to each feature independently. By contrast, for ZeMA and STRATH, we obtain the min-max values from the train set for each sensor independently, instead of each feature, to retain the shape of signal. Note that we prevent train-test bias when by fitting the scaler to the train set only instead of the entire dataset.

\section{Additional results} \label{app:res-uncood}

\begin{figure}[htbp]
	\centering
		\begin{subfigure}[H]{\textwidth}
	\caption{ODDS}
    \includegraphics[width=\textwidth]{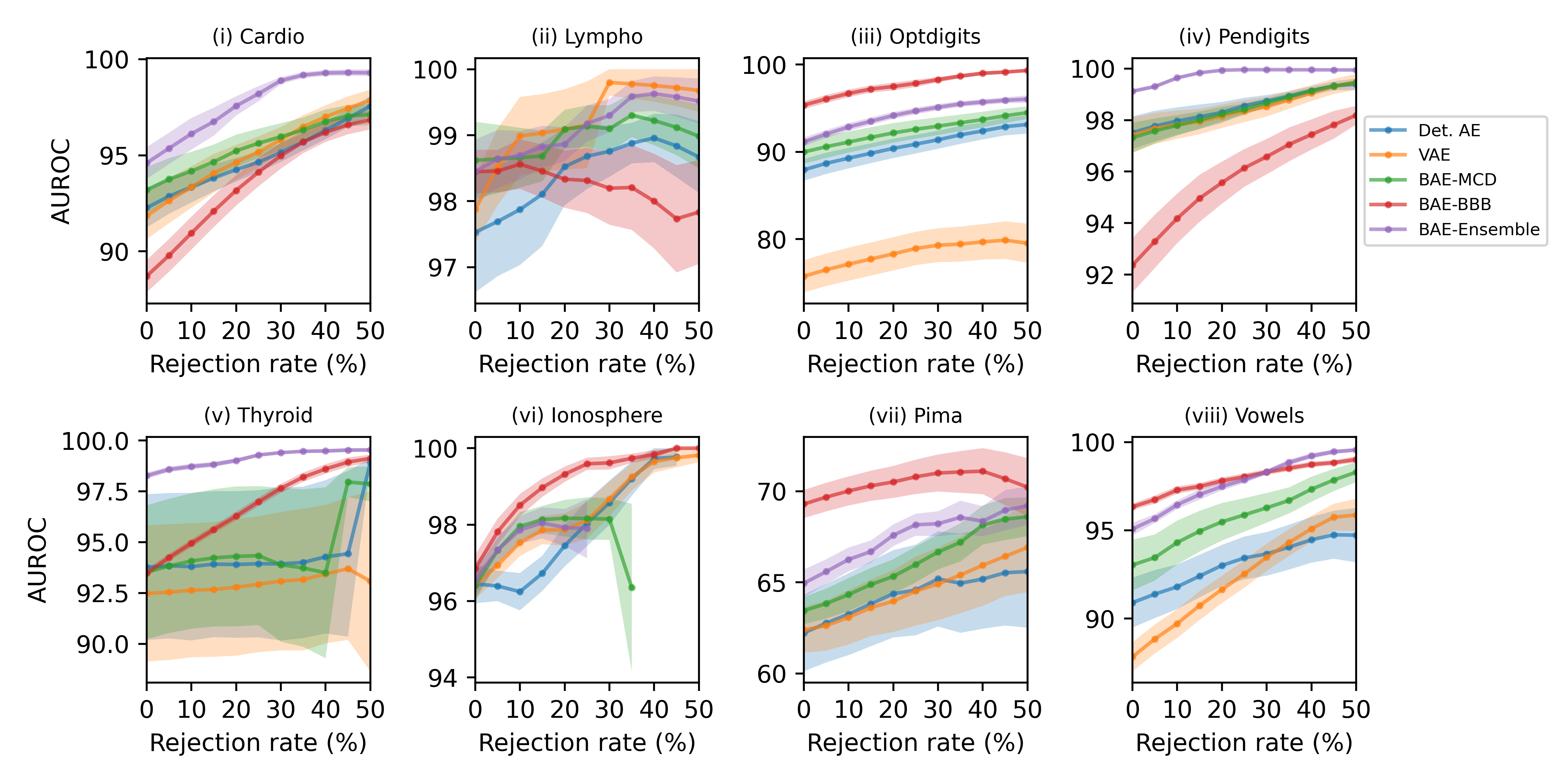}
	\end{subfigure}
	\quad
	\begin{subfigure}[H]{\textwidth}
	\caption{ZeMA}
    \includegraphics[width=\textwidth]{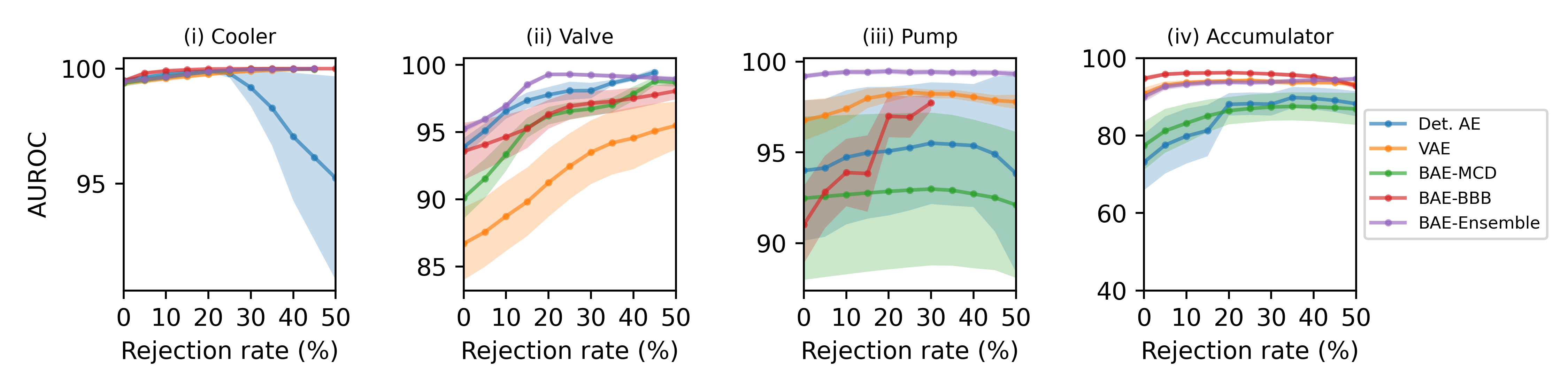}
	\end{subfigure}
	\quad
	\begin{subfigure}[H]{\textwidth}
	\caption{STRATH}
    \includegraphics[width=\textwidth]{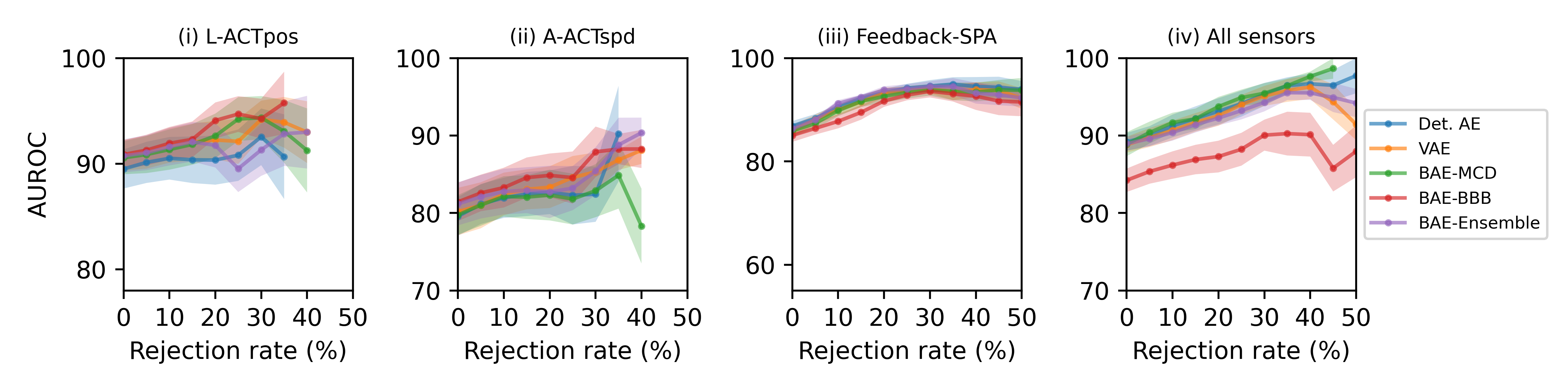}
	\end{subfigure}
	\caption{ARCs for comparisons of deterministic AE, VAE and BAEs with posterior approximated under various techniques. Mean and standard error of AUROC are evaluated on (a) ODDS, (b) ZeMA and (c) STRATH datasets over 10 experiment runs. \utotal is used as the rejection criterion with the Q distribution chosen based on the best \WAUROC score.} \label{fig:app-arc}
\end{figure}

\begin{figure}[htbp]
	\centering
	\begin{subfigure}[t]{\textwidth}
	\caption{ODDS}\label{fig:arc-dist-a}
    \centerline{\includegraphics[scale=.55]{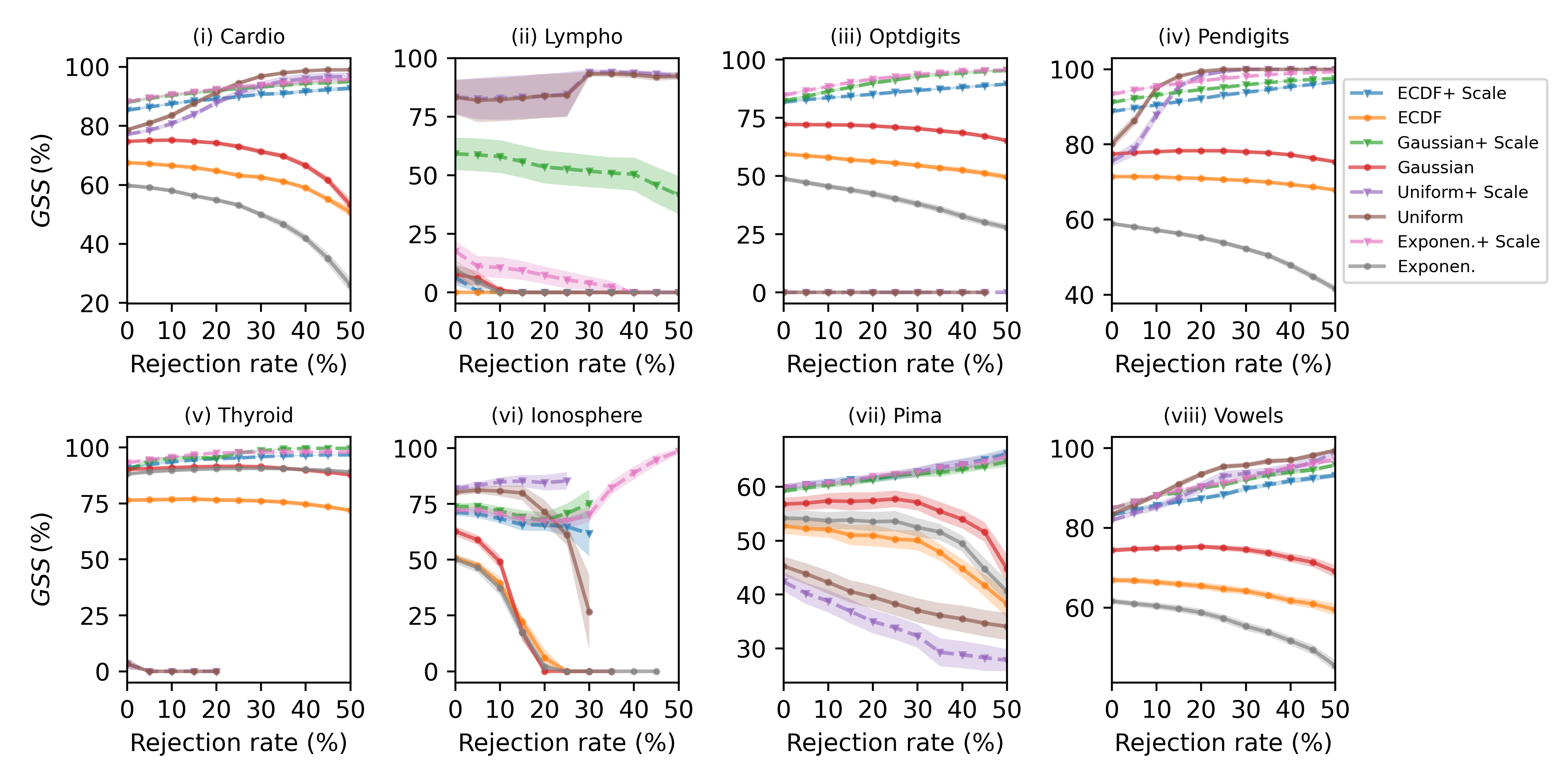}}
	\end{subfigure}
	\quad
	\begin{subfigure}[t]{\textwidth}
	\caption{ZeMA}\label{fig:arc-dist-b}
    \centerline{\includegraphics[scale=.55]{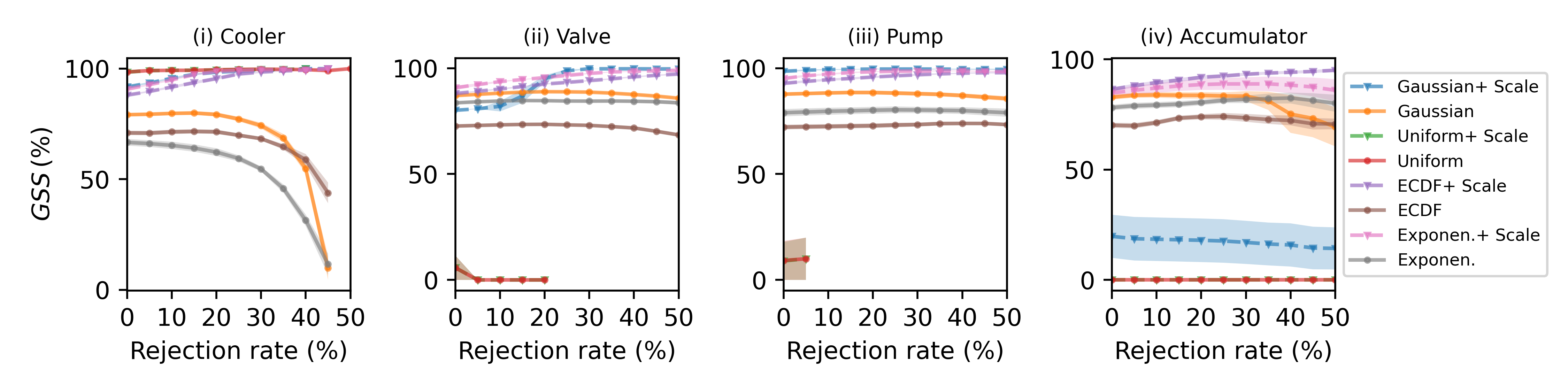}}
	\end{subfigure}
	\quad
	\begin{subfigure}[t]{\textwidth}
	\caption{STRATH}\label{fig:arc-dist-c}
    \centerline{\includegraphics[scale=.55]{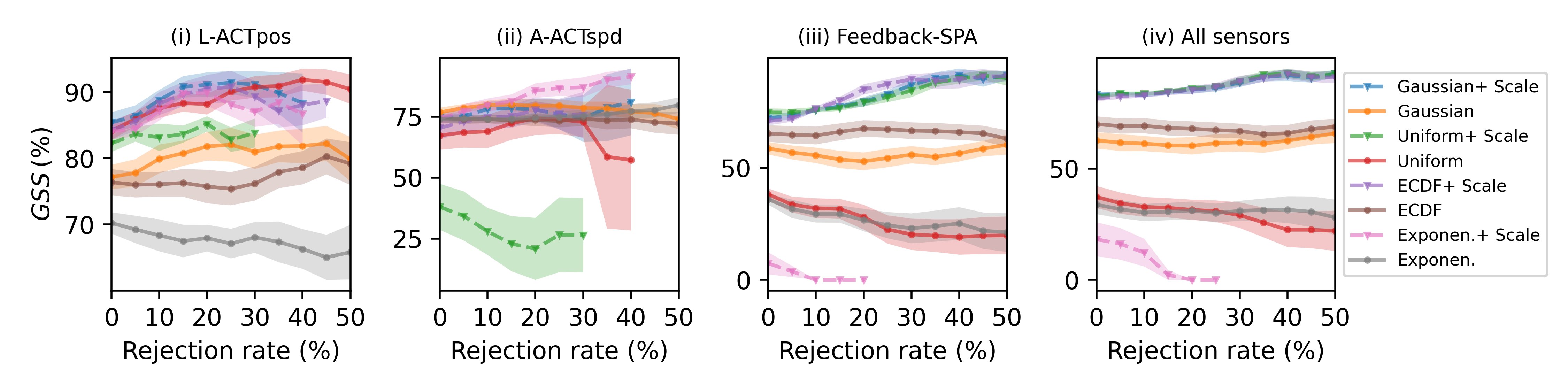}}
	\end{subfigure}
	\caption{ARCs for comparisons of anomaly probability conversions. Mean and standard error of GSS are evaluated on (a) ODDS, (b), ZeMA and (c) STRATH datasets over 10 experiment runs. Results are shown for the BAE-Ensemble model using \utotal as the rejection criterion.} \label{fig:app-proba}
\end{figure}

Results of using the area under the receiver operating characteristic curve (AUROC) \cite{bewick2004statistics} as a measure of classifier performance for ARCs are reported in \cref{fig:app-arc} and \cref{table:app-wauroc}. In addition, we report the ARCs to compare the choices of distribution Q (see \cref{fig:app-proba}).  

\begin{table}[H] 
\caption{Mean \WAUROC ($Gain^{AUROC}$) is evaluated on (a) ODDS, (b) ZeMA and (c) STRATH datasets. Each dataset consists of multiple tasks and 10 experiment runs. Uncertainty estimation method with the highest mean \WAUROC score is bolded. Values are shown in percentage.}
    \begin{subtable}[h]{\textwidth}
        \centering
        \scriptsize \caption{ODDS}
            \resizebox{\textwidth}{!}{
                \begin{tabular}{@{}llllll@{}}
                \toprule
                Model & \uepi & \ualea & \utotal & \uexceed & \VarNLL \\ \midrule
Deterministic AE & - & 91.6(+1.8) & 91.6(+1.8) & 90.1(+0.2) & - \\
VAE & 89.0(+1.2) & 89.8(+2.0) & 90.0(+2.2) & 88.0(+0.2) & 86.4(-1.3) \\
BAE-MCD & 91.5(+0.8) & 92.6(+1.9) & 92.6(+2.0) & 90.5(-0.2) & 88.3(-2.4) \\
BAE-BBB & 92.8(+1.4) & 93.5(+2.1) & 93.9(+2.5) & 91.6(+0.3) & 89.7(-1.7) \\
BAE-Ensemble & 93.2(+0.9) & 93.8(+1.5) & 94.3(+2.0) & 92.4(+0.2) & 90.2(-2.1) \\ \midrule
\textbf{Mean} & 91.6(+1.1) & 92.3(+1.9) & \textbf{92.5(+2.1)} & 90.5(+0.1) & 88.6(-1.9) \\ \bottomrule
                \end{tabular}
            }
    \end{subtable}
    \hfill
    
    \begin{subtable}[h]{\textwidth}
        \centering
        \vskip 0.2in
        \scriptsize \caption{ZeMA}
            \resizebox{\textwidth}{!}{
                \begin{tabular}{@{}llllll@{}}
                \toprule
                Model & \uepi & \ualea & \utotal & \uexceed & \VarNLL \\ \midrule
Deterministic AE & - & 92.7(+2.6) & 92.7(+2.6) & 90.8(+0.6) & - \\
VAE & 94.7(+1.3) & 95.7(+2.3) & 95.8(+2.4) & 94.0(+0.5) & 94.9(+1.4) \\
BAE-MCD & 91.7(+1.8) & 92.8(+3.0) & 92.9(+3.0) & 90.4(+0.6) & 92.2(+2.4) \\
BAE-BBB & 96.0(+1.4) & 96.3(+1.6) & 96.5(+1.8) & 95.3(+0.6) & 95.1(+0.4) \\
BAE-Ensemble & 97.5(+1.5) & 97.5(+1.6) & 97.8(+1.8) & 96.2(+0.3) & 97.0(+1.0) \\ \midrule
\textbf{Mean} & 95.0(+1.5) & 95.0(+2.2) & \textbf{95.1(+2.3)} & 93.3(+0.5) & 94.8(+1.3) \\ \bottomrule
                \end{tabular}
            }
    \end{subtable}
    \hfill

    \begin{subtable}[h]{\textwidth}
        \centering
        \vskip 0.2in
        \scriptsize \caption{STRATH}
            \resizebox{\textwidth}{!}{
                \begin{tabular}{@{}llllll@{}}
                \toprule
                Model & \uepi & \ualea & \utotal & \uexceed & \VarNLL \\ \midrule
Deterministic AE & - & 89.4(+3.1) & 89.4(+3.1) & 88.5(+2.2) & - \\
VAE & 89.4(+2.9) & 89.6(+3.1) & 89.6(+3.1) & 88.7(+2.2) & 88.2(+1.7) \\
BAE-MCD & 89.3(+3.0) & 89.5(+3.2) & 89.4(+3.1) & 88.2(+1.9) & 86.9(+0.6) \\
BAE-BBB & 88.9(+3.5) & 88.6(+3.2) & 88.4(+3.1) & 87.6(+2.2) & 86.8(+1.5) \\
BAE-Ensemble & 89.1(+2.3) & 89.8(+3.0) & 89.8(+3.0) & 89.0(+2.2) & 87.7(+0.9) \\ \midrule
\textbf{Mean} & 89.2(+2.9) & \textbf{89.4(+3.1)} & 89.3(+3.1) & 88.4(+2.1) & 87.4(+1.2) \\ \bottomrule
                \end{tabular}
            }
       
     \end{subtable}
     \label{table:app-wauroc}
\end{table}

\end{document}